\documentclass[conference]{IEEEtran}
\IEEEoverridecommandlockouts

\usepackage{cite}
\usepackage{amsmath,amssymb,amsfonts}
\usepackage{algorithmic}
\usepackage{graphicx}
\usepackage{textcomp}
\usepackage{xcolor}
\usepackage{float}
\usepackage{booktabs}

\def\BibTeX{{\rm B\kern-.05em{\sc i\kern-.025em b}\kern-.08em
    T\kern-.1667em\lower.7ex\hbox{E}\kern-.125emX}}
\begin{document}

\title{Fully Explainable Classification Models Using Hyperblocks}

\author{\IEEEauthorblockN{Austin Snyder\textsuperscript{*}}
\IEEEauthorblockA{\textit{Computer Science Department} \\
\textit{Central Washington University}\\
Ellensburg, United States \\
0009-0004-0473-6688}
\and
\IEEEauthorblockN{Ryan Gallagher\textsuperscript{*}}
\IEEEauthorblockA{\textit{Computer Science Department} \\
\textit{Central Washington University}\\
Ellensburg, United States \\
0009-0002-1928-8609}
\and
\IEEEauthorblockN{Boris Kovalerchuk}
\IEEEauthorblockA{\textit{Computer Science Department} \\
\textit{Central Washington University}\\
Ellensburg, United States \\
0000-0002-0995-9539}
}

\maketitle
\begingroup
\renewcommand\thefootnote{*}
\footnotetext{These authors contributed equally to this work.}
\endgroup

\begin{abstract}
Building on existing work with Hyperblocks, which classify data using minimum and maximum bounds for each attribute, we focus on enhancing interpretability, decreasing training time, and reducing model complexity without sacrificing accuracy. This system allows subject matter experts (SMEs) to directly inspect and understand the model’s decision logic without requiring extensive machine learning expertise. To reduce Hyperblock complexity while retaining performance, we introduce a suite of algorithms for Hyperblock simplification. These include removing redundant attributes, removing redundant blocks through overlap analysis, and creating disjunctive units. These methods eliminate unnecessary parameters, dramatically reducing model size without harming classification power. We increase robustness by introducing an interpretable fallback mechanism using k-Nearest Neighbor (k-NN) classifiers for points not covered by any block, ensuring complete data coverage while preserving model transparency. Our results demonstrate that interpretable models can scale to high-dimensional, large-volume datasets while maintaining competitive accuracy. On benchmark datasets such as WBC (9-D), we achieve strong predictive performance with significantly reduced complexity. On MNIST (784-D), our method continues to improve through tuning and simplification, showing promise as a transparent alternative to black-box models in domains where trust, clarity, and control are crucial.
\end{abstract}

\begin{IEEEkeywords}
Hyperblocks, Explainable ML, Interpretability
\end{IEEEkeywords}

\section{Introduction}
This paper focuses on Hyperblocks (HBs) as interpretable means for solving classification tasks~\cite{hyperboxSurvey2021}. HBs define class regions using axis-aligned intervals across attributes and are inherently interpretable due to this. HBs structure allows for lossless visualization in high-dimensional space using techniques such as Parallel Coordinates (PC) and other General Line Coordinates (GLC) \cite{HuberKovalerchuk2024}. 

A HB consists of lower and upper bounds (intervals) on a subset of dataset attributes. HBs can also be described as an n-D rectangle due to the bound of min and max in each dimension. Not all attributes are required in each HB for successful classification of cases. Irrelevant or redundant dimensions can be removed entirely. The primary strength of HBs lies in their explainability. Unlike black-box models, HBs enable domain experts to understand and verify the classification logic behind each decision. Despite the simplicity of HBs the challenges are in multiplicity of HBs, the need to simplify the HBs, to minimize their number for a given dataset with high accuracy, and minimizing overfitting. The goal of this paper is to progress in addressing these challenges. We demonstrate significant improvements relative to prior work.

The remainder of this paper is organized as follows: Section II defines HBs and their generation process. Section III introduces novel simplification algorithms. Section IV presents alternative approaches. Section V covers case studies, Section VI concludes the paper and presents an outline of future work.

\section{Hyperblocks}
\subsection{Concept of Hyperblocks}
A HB is defined as a set of axis-aligned intervals, one for each attribute in the dataset, which together form a bounded hyper-rectangle. The key idea is to discover regions in the feature space that are pure, meaning the region only contains samples from a single class. When classifying a new point, if it falls fully within this bounded region, it is assigned the corresponding class label of that region, i.e., the HB. To put the classification more formally: 
\begin{align*}
\mathbf{x} &= (x_{1},x_{2},\dots,x_{n}) \in \mathbb{R}^{n}, \\[6pt]
m &= (m_{1},m_{2},\dots,m_{n}), \quad
M = (M_{1},M_{2},\dots,M_{n}), \\[6pt]
&\quad\text{with }m_{i} \le M_{i}\quad(i=1,\dots,n), \\[6pt]
H &= \bigl\{\mathbf{x}\in\mathbb{R}^{n}\;\big|\;
m_{i}\le x_{i}\le M_{i},\;\forall\,i=1,\dots,n\bigr\}, \\[6pt]
\text{Class}(\mathbf{x}) &=
\begin{cases}
L, & \text{if } \mathbf{x} \in H \\
\text{undefined}, & \text{if } \mathbf{x} \notin H
\end{cases} \\[6pt]
&\quad \text{where } L \text{ is the dominant class of points in } H.
\end{align*}

\subsection{Motivation for Hyperblocks}
The motivation for using HBs as a classification model lies in their explainable and interpretable nature. In high stakes domains such as medical diagnostics, domain experts, such as cardiologists, require justifiable and transparent logical reasoning behind predictions. Consider, for example, a model that classifies a heart disease with high accuracy using an algebraic expression such as:  
\[
\frac{0.5\,\mathrm{weight}}
     {1.3\,\mathrm{age} + 2.1\,\mathrm{smoking\ status}}
\;-\;9.04\,\mathrm{cholesterol}
\]

While this expression may achieve high performance on cross-validation or test sets, it provides no intuitive explanation of decision logic. A cardiologist without a background in machine learning or mathematics cannot evaluate whether the model's reasoning aligns with their domain-specific knowledge. As a result, the model becomes untrustworthy not due to its performance metrics, but because its internal logic is inaccessible to the experts best positioned to assess its validity. The goal of HBs are to provide state-of-the-art accuracy, while being easily interpretable by a domain expert with no training in ML.

Many widely known and used classifiers, such as Linear Discriminant Functions (LDF), Support Vector Machines (SVM), and k-Nearest Neighbors (k-NN), suffer from limited interpretability. These methods often compress heterogeneous attributes into abstract numeric n-D space, which obscures the semantic difference and heterogeneity of the original features. 

In the case of k-NN, standard distance metrics such as Euclidean, Manhattan, or Mahalanobis, reduce multidimensional differences into a single scalar value, which hides the contribution of individual attributes. This transformation to a scalar value can misrepresent the true significance of individual attributes. For example, a large difference in an attribute that holds little importance to a SME can still influence the distance calculation, potentially overriding more relevant attributes. As a result, these models may misclassify critical cases and fail to earn the trust of experts in high-stakes domains. 

\subsection{Hyperblocks Generation}
Creating efficient HBs in a high-dimensional space with a large number of n-D points is quite computationally intense. First we summarize the IntervalHyper algorithm, initially developed in \cite{KovalerchukHayes2021}.  

The purpose of the IntervalHyper algorithm is to create basic HBs from pure single dimensional intervals, in turn reducing the task of merging together all combinations of n-D points to HBs. It is a greedy algorithm to create an initial set of HBs and does not guarantee an optimal set of initial HBs. This algorithm serves as a first step. Later, in the CUDA Merger Hyper (CMH) algorithm, we will take these simple HBs, and attempt to merge their bounds to reduce the complexity of the model, and find dominant patterns in the data.  
 
Our modification of the IntervalHyper algorithm from \cite{lincolnThesis} puts every point into an HB, even if it the HB only contains the single n-D point. However, if these single-point HBs remain unmerged by the end of the generation process, they are removed from the final HB set.

\subsection{Generation of Joint Hyperblocks}
After executing the IntervalHyper Algorithm, the HBs created are merged by the CUDA Merger Hyper (CMH) algorithm. The CMH algorithm aims to combine two HBs of the same class together into a single, larger HB. This is achieved by expanding each attribute's interval to cover the full extent of both input blocks: for each dimension, the new HB takes the lower of the two minimum bounds and the higher of the two maximum bounds. Once the merged bounds are constructed, a validation step is performed to ensure that no training point from an incorrect class lies within the expanded n-D region. This prevents the formation of impure or overly generalized HBs that could compromise classification accuracy. 
The \textbf{CMH} algorithm more formally is as follows as this series of steps:
\begin{enumerate}
  \item Select a seed HB from the HBs from the current class.
  
  \item For each other HB in the same class:
  \begin{itemize}
    \item Compute merged bounds by taking the element-wise minimum of lower bounds and maximum of upper bounds across all dimensions.
  \end{itemize}
  
  \item Check if any training point from a different class lies within all intervals of the merged bounds.
  
  \item If no opposing-class points are found:
  \begin{itemize}
    \item Mark the seed HB for removal.
    \item Update the partner HB to use the merged bounds.
    \item \textit{Note: Only one successful merge is required to discard the seed.}
  \end{itemize}
  
  \item If any opposing-class point is inside the merged bounds:
  \begin{itemize}
    \item Discard the merge. Leave both HBs unchanged.
  \end{itemize}
  
  \item Repeat with the next unprocessed HB as the new seed.
  \begin{itemize}
    \item Skip HBs already used as seed blocks entirely.
  \end{itemize}
\end{enumerate}

\begin{figure}[H]
    \centering
    \includegraphics[width=0.70\linewidth]{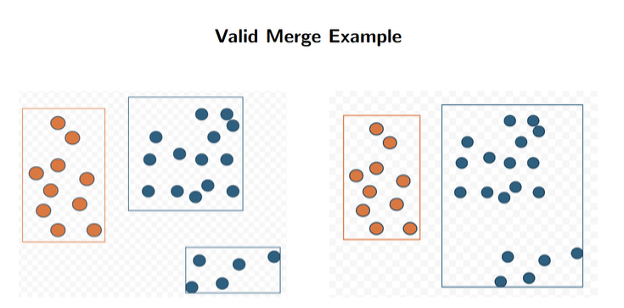}
    \caption{Example of a valid Hyperblock Merge in 2D.}
    \label{fig:validMerge}
\end{figure}

\begin{figure}[H]
    \centering
    \includegraphics[width=0.70\linewidth]{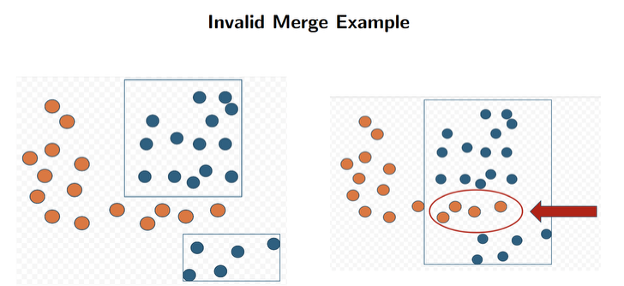}
    \caption{Example of an invalid Hyperblock Merge in 2D.}
    \label{fig:invalidMerge}
\end{figure}

Figures~\ref{fig:validMerge} and~\ref{fig:invalidMerge} display a simplified example of CMH. These show the issue addressed by the CMH algorithm: ensuring that no points from an opposing clases fall within the merged region. In Fig.~\ref{fig:validMerge}, the two blocks are successfully merged since orange points lie between the blue HBs. In contrast, Fig.~\ref{fig:invalidMerge} shows a failed merge, where points from the orange class lie between the two blue HBs.

There are a few considerations when choosing a seed HB. We keep a queue to determine which HB becomes the next seed HB. This queue does not iterate through the HBs directly however. After each seed merging iteration, if a HB merges to the seed HB, we move that HB towards the end of the queue.  

This is done by sorting the HBs in the seed queue based on whether or not they have merged on the previous iteration. With the special consideration that we move HBs which started this iteration further forward, further towards the end compared to other HBs which have merged. This helps the larger HBs have a chance to absorb more HBs, and therefore points, before their turn as the seed block.  

At the end of the CMH algorithm, we will have established a set of HBs, which can’t be merged with one another any further. The only HBs which are considered further past this algorithm are those which were not able to be merged when they were the seed HB.

\section{Simplifications of Hyperblocks}
To reduce model complexity and improve interpretability, we propose a suite of simplification methods. The primary goals are to minimize the total number of clauses and blocks while mitigating overfitting, particularly the kind that emerges from generating many small, overly specific HBs. We introduce three novel simplification techniques: Remove Redundant Attributes (R2A), Disjunctive-Unit HBs, and Remove Redundant Blocks (R2B). Each of these methods targets a different aspect of complexity within the model and collectively contribute to a more compact and generalized rule set.

\subsection{Removing Redundant Attributes}
The first proposed method for model simplification is the Remove Redundant Attributes (R2A) algorithm. This technique reduces model complexity by identifying and eliminating individual attribute constraints within a HB that do not contribute to its discriminative power. Attributes are removed only when their elimination does not allow any opposing-class training points into the block, thus preserving classification accuracy while reducing the number of clauses. 

An example of the benefits of this algorithm will be discussed later. In our Wisconsin Breast Cancer case studies, the largest HB of the benign class required only four out of nine clauses to classify over 400 points.

Given a set of HBs and a normalized dataset (all attributes between [0.0,1.0] for all data), this algorithm identifies and removes attribute bounds from each HB if doing so does not allow misclassification of points from opposing classes. A high-level outline of the \textbf{R2A algorithm} is as follows:
\begin{enumerate}
  \item For each HB:
  \begin{enumerate}
    \item For each attribute:
    \begin{itemize}
      \item Temporarily expand the HB's interval on the current attribute to the full \([0, 1]\) range.
      \item Check if any points from opposing classes now fall within the updated HB.
      \begin{itemize}
        \item If no opposing points are within the updated HB, keep the expanded interval.
        
        \item If opposing points are now within the updated HB, revert the HBs interval to its previous range.
      \end{itemize}
    \end{itemize}
  \end{enumerate}
\end{enumerate}

When this algorithm is done, we will have removed many attributes from the HBs. Note this is a greedy algorithm and does not guarantee the optimality of removing attributes. A domain expert may choose a better order to remove the attributes in. In our testing, we sorted the attributes by their coefficients found in a multiclass Linear Discriminant Function (LDF). The LDF is not used for classification and does not affect the definition or logic of HBs. It serves as a heuristic to optimize attribute-wise interval generalization order. 

\begin{figure}[H]
    \centering
    \includegraphics[width=0.75\linewidth]{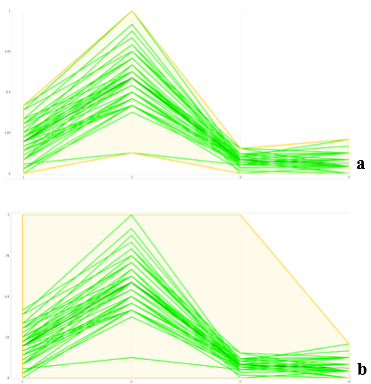}
\caption{A HB containing all 50 cases of the Iris Setosa class, shown before and after R2A simplification. Visualized using Parallel Coordinates in the DV2.0 program originally developed by \cite{lincolnThesis}.}

    \label{fig:SetosaPic}
\end{figure}

A clear example of effective attribute reduction occurs in the Fisher Iris dataset. The HB shown in Figure~\ref{fig:SetosaPic}a successfully classifies all 50 instances of the Setosa class, both before and after redundant attribute removal. The simplified version achieves 100\% coverage while using only one out of four clauses. It is shown as the leftmost attribute in Figure~\ref{fig:SetosaPic}b, where only this attribute has it's upper bound less than max value.

\subsection{Removing Redundant Hyperblocks}

The second simplification method proposed is the Remove Redundant Blocks (R2B) algorithm. This method aims to eliminate HBs that do not uniquely contribute to the classification of any training data. It is based on the idea that overly specific blocks often represent redundant or overlapping decision regions, which may lead to model overfitting. By analyzing which blocks are most general, measured by how many data points they cover uniquely, we can safely discard blocks that are unused or overshadowed by other HBs.

The \textbf{R2B algorithm} can be explained as follows:

\begin{enumerate}
  \item Take each point in the training data, and flag it as belonging to the first HB which it is inside of. (Note that a point can belong to multiple HBs, but that is handled later.)
  \item Take a count of how many points belong to each HB.
  \item Take each point again, this time flagging it belonging to the largest HB it is inside of (by count from step 2).
  \item Recount the number of cases in each HB. 
  \item Delete the HBs which have a size (count of training cases) under the removal threshold. 
\end{enumerate}

This algorithm has been shown to greatly reduce our overfitting in testing, especially when paired with the R2A (Remove Redundant Attributes) algorithm. It is often the case that even our largest HBs, classifying huge portions of the dataset, are classifying largely the same group of points. With very slight changes between them. In this case, we would have multiple HBs which do not help to actually classify any new points, since their points can be classified by other HBs.  

\subsection{Generating Disjunctive Units From Hyperblocks}

The Disjunctive HB Simplification algorithm aims to reduce the number of HBs by merging blocks of the same class when it is safe to do so. Specifically, it attempts to combine two HBs into a single, more general "disjunctive" block that represents multiple allowed intervals per attribute (i.e., logical OR conditions). A merge is permitted only if the resulting block does not include any training points from opposing classes. This strategy enables the model to consolidate overlapping or adjacent decision regions, reducing clause redundancy while preserving class purity and interpretability. 

Consider two HBs \(B_1\) and \(B_2\), both belonging to the same class. Let each block define bounds over attributes \(x_1, x_2, \ldots, x_n\). Suppose that for all \(i \neq j\), the bounds on attribute \(x_i\) are identical in both blocks:
\[
[\min_{B_1}^{(i)},\, \max_{B_1}^{(i)}] = [\min_{B_2}^{(i)},\, \max_{B_2}^{(i)}].
\]
For attribute \(x_j\), assume the bounds are non-overlapping and distinct:
\[
[\min_{B_1}^{(j)},\, \max_{B_1}^{(j)}] \cap [\min_{B_2}^{(j)},\, \max_{B_2}^{(j)}] = \emptyset.
\]
If merging the bounds on \(x_j\) into a disjunctive set,
\[
[\min_{B_1}^{(j)},\, \max_{B_1}^{(j)}] \cup [\min_{B_2}^{(j)},\, \max_{B_2}^{(j)}],
\]
results in a HB that does not include any points from other classes, then \(B_1\) and \(B_2\) may be merged into a single disjunctive HB. In the generalized case, any number of such disjunctions may be applied, provided the resulting HB preserves class purity.

\begin{figure}[H]
    \centering
    \includegraphics[width=0.75\linewidth]{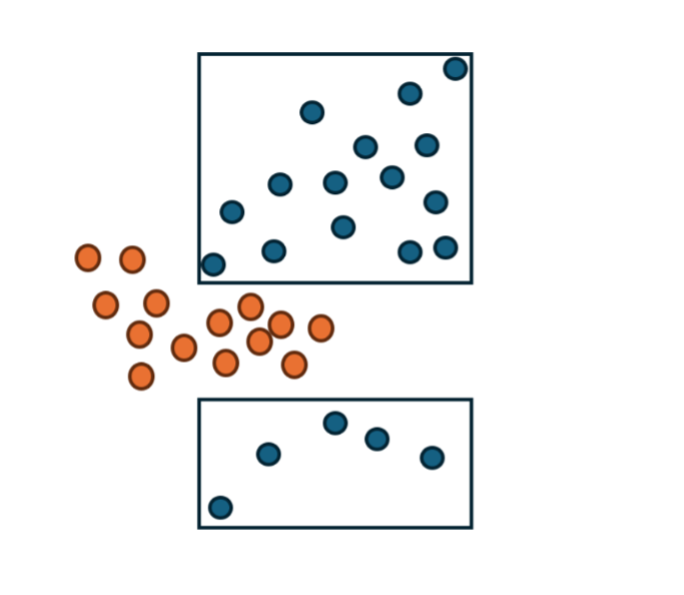}
    \caption{2D Disjunctive Unit (Disjunctive HB).}
    \label{fig:disjunction}
\end{figure}

Fig.~\ref{fig:disjunction} presents a 2D example of a disjunctive unit. In this case, two HBs share the same width interval, while differing in height intervals. By introducing a disjunction, we allow a point to fall within either HBs height interval. This avoids duplicating the shared width. In higher dimensions, such as with nine attributes, applying a disjunction to just one attribute enables the combination of two HBs using only ten clauses instead of the original eighteen, improving model compactness.

\section{Alternative Approaches}
In testing, HBs generated on training data are often not able to classify all testing data. In such a case, we aim to maintain model interpretability, while using fall-back classifiers to classify points not covered by HBs. In testing, the following options have been attempted:
\begin{itemize}
    \item Taking a Manhattan or Euclidean distance to the closest bound of each attribute of all HBs. Then using the closest HB as the classification.
    \item Performing a k-NN classifier with either of these distance measures, using the k-Nearest HBs and holding a vote.
    \item k-NN with a Euclidean distance, comparing to training points.
\end{itemize}

In our experiments, none of these methods been standout performers in terms of accuracy, they also hurt interpretability. All of them represent the closeness of a new case to the HB by a single number, therefore truncating important information.

The best performing, and most interpretable method in the case where HBs do not cover a testing point, is an implementation of \cite{lincolnThesis}'s \textbf{Explainable Threshold Similarity}, and can be explained as follows: 
\begin{itemize}
    \item Form a set of thresholds \{T\}, where $T_1$, $T_2$,\dots $T_n$ correspond to each attribute of the dataset. 
    \item For each case in the training data:
    \begin{itemize}
        \item Count how many attributes $x_i$ are within $T_i$. 
    \end{itemize}
    \item Using Explainable Threshold Similarity, we run a \textit{k}NN, where our distance measure is the similarity score, and we hold a simple majority vote with the K most similar points.
\end{itemize}
\enlargethispage{2\baselineskip}
Defining an effective set of thresholds is a nontrivial task. An SME can draw on domain knowledge to set thresholds that reflect meaningful distinctions between values. Lacking this experience, we instead define thresholds as a fixed fraction of the standard deviation for each attribute. By default, all thresholds are equal, though tuning or expert input has been seen to improve performance.

\section{Case Studies}
\subsection{Wisconsin Breast Cancer Diagnostic}

Our first case study is the Wisconsin Breast Cancer dataset. We are using the 9 dimensional, 683 cases, 2 class dataset. The dataset has 444 benign cases, and 239 malignant cases.

\begin{figure}
    \centering
    \includegraphics[width=0.75\linewidth]{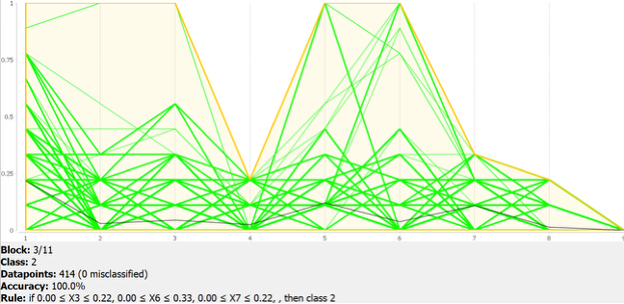}
    \caption{Largest HB of Benign class for WBC.}
    \label{fig:wbchealthy}
\end{figure}

\begin{figure}
    \centering
    \includegraphics[width=0.75\linewidth]{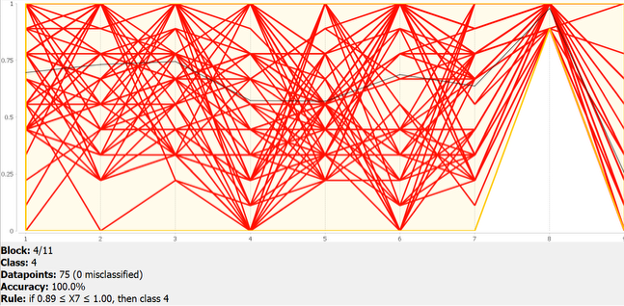}
    \caption{Largest HB of Malignant class for WBC.}
    \label{fig:wbccancer}
\end{figure}

Through 10 Fold Cross Validation, we found an average accuracy of 96.6\%. Achieved with our simplified HBs, followed by a KNN based on our Explainable Threshold Similarity scores. This accuracy is competitive with uninterpretable models such as MLP, Random Forest, and SVM as seen in Table ~\ref{tab:wbc_othermodels} from \cite{lincolnThesis} augmented with our result in bold. This is done while maintaining explainability for SME analysis. Not only is our classification explainable, it is losslessly visualizable as well. Displayed in Figure~\ref{fig:wbchealthy} and Figure~\ref{fig:wbccancer} are the largest blocks of each class in the dataset, representing 93.2\% of the benign cases, and 31.3\% of malignant cases respectively. Notably, the HB for the benign class classifies all of the cases with only four clauses. Since only \(x_3, x_6, x_7,\) and \(x_9\) take values other than the minimum \(0.0\) or maximum \(1.0\), we consider only the corresponding rules.

\begin{table}[H]
  \centering
  \caption{Average 10-fold Accuracy for WBC}
  \label{tab:wbc_othermodels}
  \begin{tabular}{l c}
    \toprule
    Model           & Avg.\ Accuracy (\%) \\
    \midrule
    Decision Tree (DT)     & 93.53 \\
    Support Vector Machine & 96.33 \\
    k-Nearest Neighbors    & 96.91 \\
    Linear Discriminant Analysis & 95.44 \\
    Multi-Layer Perceptron & 96.77 \\
    Random Forest (RF)     & 96.91 \\
    \textbf{Simplified Hyperblocks} & \textbf{96.66\%} \\
    \bottomrule
  \end{tabular}
\end{table}

Additionally, only one clause, on attribute $x_7$ , is needed to classify 31.38\% of the malignant cases into the HB.

The per-fold accuracies, block count, and clause counts are displayed in Table~\ref{tab:wbc_folds} and averages in Table ~\ref{tab:wbc_stats}. In combination with Explainable Threshold Similarity (ETS) based k-NN, we achieve 100\% classification coverage. The parameters used are as follows: removal threshold = 5, k = 5, and ETS thresholds $T_i$ where:
$T_i$ = 0.25 * std dev $x_i$ for all attributes $x_i$. We attempted to optimize these parameters through grid search, however, only minor improvements were found.

\begin{table}[H]
\centering
\caption{Fold‐by‐fold results for accuracy, block count, and clause count.}
\begin{tabular}{ccc}
\toprule
Accuracy \% & Block Count & Total Clauses \\
\midrule
98.55 &  9 & 23 \\
98.55 & 11 & 34 \\
97.10 & 10 & 31 \\
98.55 &  9 & 28 \\
97.05 & 11 & 33 \\
97.05 &  9 & 23 \\
92.65 & 10 & 29 \\
 100.00 & 10 & 30 \\
94.11 &  9 & 28 \\
92.54 & 10 & 25 \\
\bottomrule
\end{tabular}
\label{tab:wbc_folds}
\end{table}

\begin{table}[H]
\centering
  \caption{Some basic statistics for these 10 folds of results.}
  \label{tab:wbc_stats}
\begin{tabular}{lccc}
\toprule
Statistic             & Accuracy \% & Block Count & Clause Count \\
\midrule
Average               &  96.62      &       9.8    &        28.4   \\
Standard Deviation    &   2.49      &       0.75   &         3.64  \\
Minimum               &  92.54      &       9      &        23     \\
Maximum               & 100.00      &      11      &        34     \\
\bottomrule
\end{tabular}
\end{table}
 
 In the case study done in \cite{lincolnThesis}, HBs achieved a 10-fold cross validation accuracy of 97.74\%. Our simplified model uses just 9.7 HBs on average, with a total of 27.1 clauses on average. As compared with the former’s 37.7 HBs on average, each with all 9 attributes, for a total of 339.3 clauses. Simplified HBs maintain competitive accuracy at only 1.12\% lower than \cite{lincolnThesis}, while reducing complexity by 92.1\%. Considering the small size of the WBC dataset, the accuracy lost represents at most 2-3 more misclassifications. 

The reduction in clauses increases interpretability by reducing the number of rules needed to be analyzed. The simplicity coupled with competitive accuracy, displays that HBs are a powerful tool to be used in high stakes domains such as cancer classification.

\subsection{MNIST: Digits 2 and 7}

In this study, our model is evaluated using a binary classification task restricted to digits 2 and 7 from the MNIST hand-drawn digit dataset. This subset was chosen to allow for comparison with \cite{lincolnThesis}, who used digits 2 and 7 to benchmark rule-based explainable models due to their high visual similarity and classification difficulty. We use the Kaggle-hosted “MNIST in CSV” \cite{DatoOn2018}, which provides the dataset in a flattened, 784-dimensional, grayscale CSV format. All digit 2 and 7 samples from the 60,000-record training set are extracted and are used in 10-fold cross-validation. The 10,000-record MNIST test set is not used in this study.

Our adapted HB generation approach significantly improves model efficiency, while maintaining competitive accuracy compared to previous work on HB-based MNIST classification. In \cite{lincolnThesis}, a two-level HB generation framework previously proposed in \cite{HuberKovalerchuk2024}, was applied to distinguish digits 2 and 7 from the MNIST dataset. The Level 1 k-NN HBs averaged 304.2 HBs, which were reduced by 34\% to 200.6 at Level 2 \cite{lincolnThesis}, as shown in Table~\ref{tab:hb-counts}. This refinement increased their classification accuracy from 98.31\% to 99.44\%, as shown in the first two columns of Table \ref{tab:accuracy-results} ~\cite{lincolnThesis}. The corresponding accuracy values for our unsimplified and simplified HBs across 10 folds are presented in the last two columns of the same table.

\begin{table}[htbp]
\centering
\caption{Fold-wise HB Counts}
\label{tab:hb-counts}
\begin{tabular}{c|c|c|c|c}
\hline
\textbf{Fold} & \textbf{L1} & \textbf{L2} & \textbf{Unsimp.} & \textbf{Simp.} \\
\hline
1  & 276 & 219 & 128 & 43 \\
2  & 260 & 211 & 126 & 46 \\
3  & 328 & 223 & 130 & 40 \\
4  & 337 & 188 & 129 & 45 \\
5  & 318 & 240 & 125 & 46 \\
6  & 273 & 190 & 129 & 42 \\
7  & 327 & 199 & 124 & 45 \\
8  & 318 & 204 & 121 & 41 \\
9  & 290 & 207 & 129 & 46 \\
10 & 315 & 205 & 130 & 46 \\
\hline
\textbf{Avg} & 304.2 & 208.6 & 127.1 & 44 \\
\hline
\end{tabular}
\end{table}

\begin{table}[htbp]
\centering
\caption{Fold-wise Accuracy (\%)}
\label{tab:accuracy-results}
\begin{tabular}{c|c|c|c|c}
\hline
\textbf{Fold} & \textbf{L1} & \textbf{L2} & \textbf{Unsimp.} & \textbf{Simp.} \\
\hline
1  & 99.08 & 99.68 & 98.77 & 99.26 \\
2  & 98.25 & 99.33 & 99.68 & 98.69 \\
3  & 99.36 & 99.61 & 98.03 & 98.93 \\
4  & 99.21 & 99.59 & 98.28 & 99.10 \\
5  & 99.37 & 99.62 & 99.01 & 99.10 \\
6  & 99.42 & 98.64 & 98.03 & 98.44 \\
7  & 99.18 & 99.76 & 98.19 & 98.85 \\
8  & 98.12 & 99.30 & 98.52 & 99.01 \\
9  & 91.33 & 99.48 & 98.03 & 98.77 \\
10 & 99.74 & 99.35 & 98.11 & 98.60 \\
\hline
\textbf{Avg} & 98.31 & 99.44 & 98.28 & 98.87 \\
\hline
\end{tabular}
\end{table}

\begin{figure}
    \centering
    \includegraphics[width=1.0\linewidth]{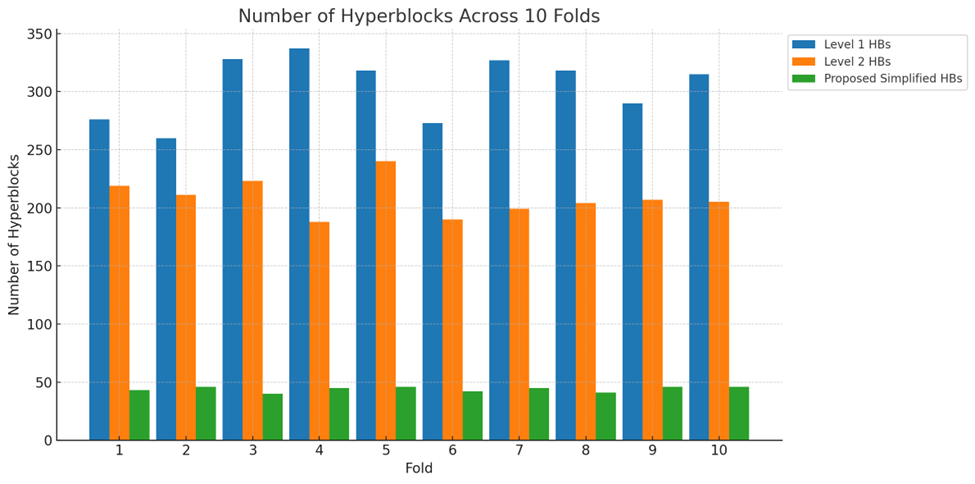}
    \caption{Comparison of Level 1 and 2 HBs to our simplified HBs in terms of block count.}
    \label{fig:foldWiseHBs}
\end{figure}

\begin{figure}
    \centering
    \includegraphics[width=1.0\linewidth]{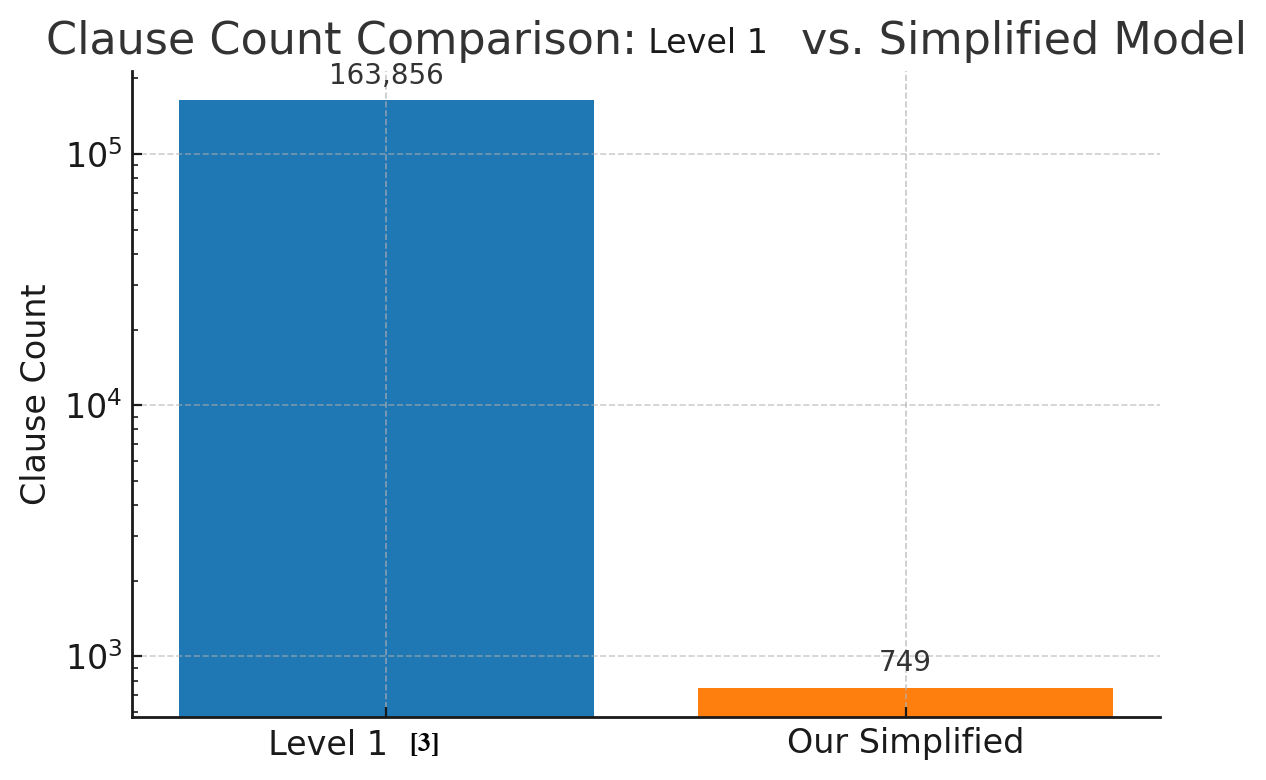}
    \caption{Comparison of Level 1 HBs \cite{lincolnThesis} to our simplified HBs in terms of clause count.}
    \label{fig:clauseCount27}
\end{figure}

With our pre-simplification HBs, we achieved an average of 127.1 HBs, representing a 58.2\% reduction vs. their Level 1 model, and a 39.1\% reduction vs. their Level 2 model. Our average number of clauses that form rules (rule count) is 76,280.70, compared to their reported 163,856 clauses, representing a 52.22\% reduction in logical rule volume. Despite these simplifications, our approach maintains an average accuracy of 98.28\%, within 0.03\% of Level 1 HBs and just 1.16\% below Level 2 HBs. After applying the simplification suite, comprising the R2A and R2B algorithms, during 10-fold cross-validation, we achieved an average of only 44 HBs per fold, as shown in Fig.~\ref{fig:foldWiseHBs}. The clause count was further reduced to an average of 749 per fold, a dramatic reduction from the initial Level 1 HBs 163,856 as seen in Fig.~\ref{fig:clauseCount27}.

These simplifications yielded an average classification accuracy of 98.87\%, outperforming both our unsimplified model (98.28\%) and \cite{lincolnThesis}’s Level 1 model (98.31\%), while approaching their Level 2 accuracy of 99.44\%, despite using over 99.54\% fewer rules and 78.9\% fewer blocks. Table \ref{tab:accuracy-results} shows fold-wise classification accuracy for \cite{lincolnThesis}’s Level 1 and Level 2 models compared to our simplified HBs. These results highlight the strength of the proposed HB generalization strategy in preserving accuracy while significantly reducing model complexity and overfitting. The substantial simplification, especially in the number of clauses, enables deep analysis of decision logic by SMEs, a critical advantage in high-stakes domains, where model transparency directly impacts trust and adoption.

\subsection{MNIST: All Digits}
This study uses the full MNIST dataset, containing 60,000 training and 10,000 test samples, across all ten digit classes. The increased scope allows us to show the scalability and competitiveness of HB based classification.

CUDA-based parallelization allowed us to significantly expand upon the work in \cite{lincolnThesis}, enabling full-dataset experimentation previously infeasible due to computational constraints. We distributed the workload across ten lab workstations (each with an RTX 4090, i9-12900 CPU, and 64 GB RAM), assigning one digit class per machine. This allowed us to generate HBs for the full training set in under 30 minutes.

Despite these gains, simplification remained computationally expensive. We therefore applied full R2B simplification and a reduced-scale R2A, targeting peripheral image regions less relevant to classification.

Table~\ref{tab:accuracy_per_digit} presents the per-digit accuracy for HBs, k-NN fallback, and the final combined result. HBs covered 70.21\% of the test set, with remaining cases handled by an explainable threshold-based k-NN. R2B simplification reduced the model from 9984 HBs and 7.83 million clauses to 2534 HBs and 1.43 million clauses, an 81.77\% reduction, while achieving 94.4\% overall accuracy.

\begin{table}[h]
\centering
\caption{Per-Digit Accuracy Comparison Across Models}
\begin{tabular}{|c|c|c|c|}
\hline
\textbf{Digit} & \textbf{HyperBlocks} & \textbf{KNN Fallback} & \textbf{Final Combined} \\
\hline
0 & 0.996 & 0.965 & 0.991 \\
1 & 0.982 & 0.976 & 0.981 \\
2 & 0.924 & 0.946 & 0.934 \\
3 & 0.953 & 0.916 & 0.942 \\
4 & 0.923 & 0.934 & 0.928 \\
5 & 0.905 & 0.961 & 0.934 \\
6 & 0.969 & 0.967 & 0.969 \\
7 & 0.962 & 0.872 & 0.942 \\
8 & 0.911 & 0.912 & 0.912 \\
9 & 0.853 & 0.952 & 0.906 \\
\hline
\textbf{Overall} & \textbf{0.948} & \textbf{0.937} & \textbf{0.944} \\
\hline
\end{tabular}
\label{tab:accuracy_per_digit}
\end{table}

LeCun’s original linear discriminant analysis (LDA) implementation for MNIST achieved an error rate of 7.6\%~\cite{lecun}, While our approach achieved an error rate of 5.56\% (accuracy of 94.44\%), surpassing this benchmark. Moreover, our approach provides significantly greater explainability: SMEs can directly trace decisions through interpretable rule-based logic and visual tools such as Parallel Coordinates, whereas LDA relies on mathematical class separation that is opaque to non-experts.

In comparison to Convolutional Neural Networks (CNNs), which achieve state-of-the-art results such as 99.77\% accuracy (error rate of 0.23\%)~\cite{ciresan}, our model does not reach that level of raw performance. However, CNNs such as VGG16 and ResNet50V2 require 138.4 million and 25.6 million parameters, respectively~\cite{keras}. In contrast, our final simplified model uses approximately 1.43 million clauses, or about 2.86 million total number values (min and max per clause). This reduction in complexity is more meaningful than it may appear: in a CNN, one could examine the learned weights for days without gaining a clear understanding of the models logic. In contrast, each decision in our simplified model can be directly traced through visualization methods such as Parallel Coordinates, providing transparent logic behind every classification.

\section{Conclusion}
This paper has shown that HBs are a viable model. The proposed methods have significantly reduced the model size, training time, and overfitting. HBs can be explained to any domain expert, and allow them to make informed decisions, while also still being competitive in accuracy with larger, uninterpretable state of the art models. We have shown that the future of Machine Learning classifiers does not have to be these black box models which cannot be fully trusted. 

\section{Future Work}

Several promising research directions remain for improving HB models. One such direction involves refining the HB voting mechanism. In its current form, classification is performed using a normalized majority vote, where each HB in a class contributes equally with weight \( \frac{1}{N} \), where \( N \) is the total number of HBs in that class. This method implicitly assumes that all HBs are equally informative, which is not always supported by empirical performance.

Experimental results suggest that some HBs have significantly higher precision and classification quality than others. To address this, we are exploring a precision-weighted voting strategy, where each HB’s contribution is scaled according to its historical precision on a held-out validation set.

Additionally, we are investigating mechanisms to redistribute a portion of an HB’s vote to other classes in proportion to its misclassification tendencies. For instance, suppose classes \( X \), \( Y \), and \( Z \) each have a single HB. If the HB for class \( X \) achieves 80\% precision on validation and misclassifies 20\% of its points as class \( Y \), then its vote can be split accordingly, assigning 0.8 weight to class \( X \) and 0.2 to class \( Y \). This adjustment may lead to more nuanced classification decisions.

Another possible task is optimization of the simplification algorithms to allow simplifications on larger, high dimensional data.

Finally, enhancing multi-class classification by leveraging combinations of pairwise binary classifiers is a promising direction. Our MNIST case study involving digits 2 and 7 demonstrated high accuracy in binary classification. Extending this level of performance to general multi-class problems via structured pairwise HB combinations could significantly improve the model's overall classification accuracy.

\nocite{*}
\bibliographystyle{IEEEtran}
\bibliography{references}

\end{document}